\documentclass[1col]{ceurart}

\usepackage{amsmath,amssymb}
\usepackage{graphicx}
\usepackage{booktabs}
\usepackage{hyperref}
\usepackage{tikz}
\usetikzlibrary{shapes.geometric, arrows.meta, positioning, fit, backgrounds}
\usepackage[ruled,vlined,linesnumbered]{algorithm2e}

\newcommand{\method}{CaptionFuse}

\begin{document}

\copyrightyear{2026}
\copyrightclause{Copyright 2026 for this paper by its authors.
  Use permitted under Creative Commons License Attribution 4.0
  International (CC BY 4.0).}

\conference{ECOM'26: SIGIR Workshop on eCommerce, Jul 24, 2026, Melbourne, Australia}

\title{CaptionFuse: Training-Free Composed Product Retrieval Via VLM Intent Captioning and Weighted Embedding Fusion}

\author[1]{Zhao Gao}[email=orangez@amazon.com]
\author[1]{Mina Metias}[email=mmetias@amazon.ae]
\author[1]{Mohamed Noordeen Alaudeen}[email=nooraldn@amazon.ae]
\author[1]{Bassel Saleh}[email=basaleh@amazon.ae]

\address[1]{AWS GenAI Innovation Center}

\begin{abstract}
Composed Image Retrieval (CIR) retrieves products matching a reference image modified by natural language---a core capability for eCommerce search.
Existing training-free methods fuse CLIP embeddings via arithmetic, but we identify a fundamental limitation: user text modifications are \emph{attribute-sparse}---a user who types ``make it red with shorter sleeves'' specifies \emph{what to change} but not \emph{what to keep} (the floral pattern, the V-neck, the fit-and-flare silhouette).
We introduce \emph{language-space fusion}: a Vision-Language Model (VLM) jointly interprets the reference image and modification to generate a \emph{visually grounded} product-title caption---e.g., ``Red floral fit-and-flare dress with short sleeves and V-neck''---that fills in the missing attributes from the image.
Our method, \textbf{\method{}}, requires no task-specific training and is model-agnostic.
A controlled experiment confirms that \emph{attribute content}---not linguistic style---drives the gains: imperative and descriptive captions with identical attributes achieve equal R@10, while both outperform raw modifications by +6.5 points.
Embedding geometry analysis shows 62\% larger discriminative margin, with 68\% of remaining failures caused by backbone limitations---not VLM errors (5\%).
On CIRCO (123K gallery), \method{} achieves 22.4\% mAP with CLIP and 31.7\% with Titan, outperforming WeiMoCIR (13.9\%), LinCIR (13.2\%), and Pic2Word (11.6\%).
On FashionIQ, \method{} matches the best training-free baseline; on CIRR, where modifications are already attribute-rich, VLM captioning provides less benefit.
We provide a detailed analysis of when and why the method succeeds or fails (\S\ref{sec:analysis}).
\end{abstract}


\maketitle

\section{Introduction}

A customer browsing an eCommerce catalog sees a blue floral dress and types ``make it red with shorter sleeves.''
This is \emph{Composed Image Retrieval} (CIR)~\cite{vo2019composing}: retrieving products that match a reference image modified by natural language.
CIR is a high-value capability for product search, yet deploying it at scale is challenging because training-based methods~\cite{delmas2022artemis,gu2024compodiff} require expensive CIR-specific triplet data that is scarce in most product catalogs.

Training-free approaches~\cite{saito2023pic2word,gu2024lincir,tang2024weimocir} avoid this cost by combining pre-trained CLIP~\cite{radford2021learning} embeddings, but they operate in \emph{embedding space} where fine-grained attribute reasoning is lost.
The core problem is that user modifications are \emph{attribute-sparse}: ``make it red with shorter sleeves'' specifies \emph{what to change} but not \emph{what to keep}---the floral pattern, the V-neck, the fit-and-flare silhouette.
Embedding arithmetic struggles with these sparse modifications for a geometric reason: CLIP's text encoder was trained on descriptive alt-text, so sparse imperative phrases (``make it red'') map to a generic ``modification intent'' region of embedding space rather than landing near specific product embeddings.
We show in \S\ref{sec:analysis} that raw modifications achieve only 0.358 mean cosine similarity to their target images, compared to 0.487 for attribute-rich VLM captions---a 36\% gap that linear combination cannot bridge when the text vector carries insufficient attribute signal.

We propose \textbf{\method{}}, which shifts fusion from embedding space to \emph{language space}.
A VLM jointly interprets the reference image and text modification, then generates a concise \emph{product-title} caption that fills in the missing attributes---e.g., ``Red floral fit-and-flare dress with short sleeves and V-neck.''
The caption is encoded and combined with the reference image embedding for nearest-neighbor retrieval.
The approach is fully training-free and model-agnostic: the VLM, embedding backbone, and vector index are all independently swappable.

\begin{figure}[t]
\centering
\includegraphics[width=\linewidth]{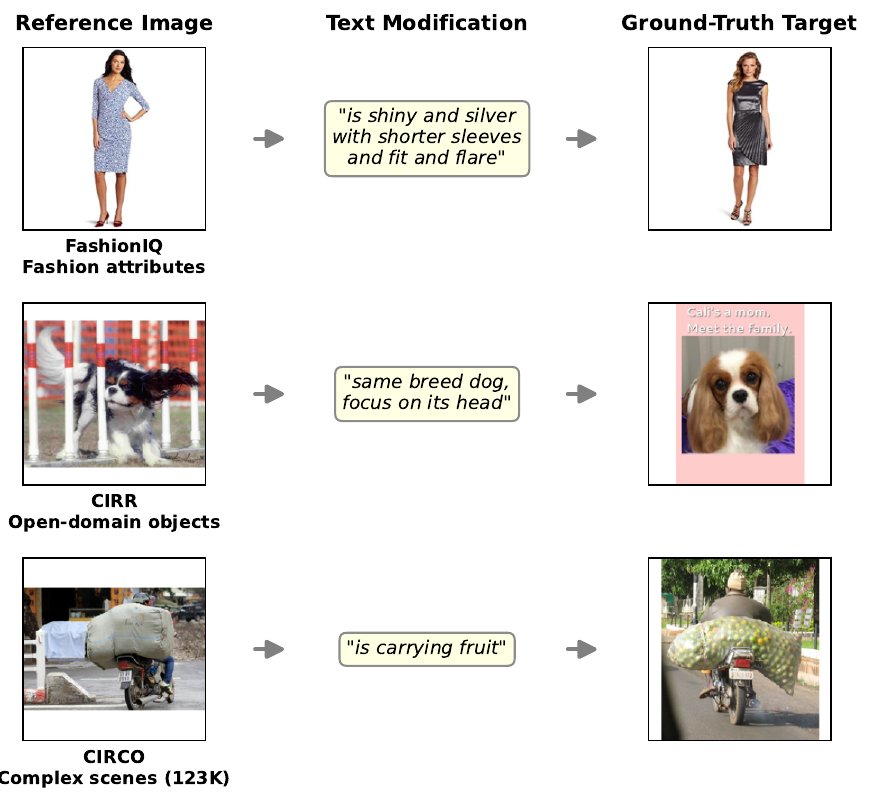}
\caption{Composed Image Retrieval examples across three benchmarks. Given a reference image and a text modification, the goal is to retrieve the target. FashionIQ tests fashion attribute changes; CIRR tests open-domain object modifications; CIRCO tests complex multi-aspect scene reasoning over a 123K image gallery.}
\label{fig:dataset_examples}
\end{figure}

Figure~\ref{fig:dataset_examples} illustrates the CIR task across our three evaluation benchmarks, highlighting the diversity of domains and modification complexity.

\begin{figure}[t]
\centering
\includegraphics[width=\linewidth]{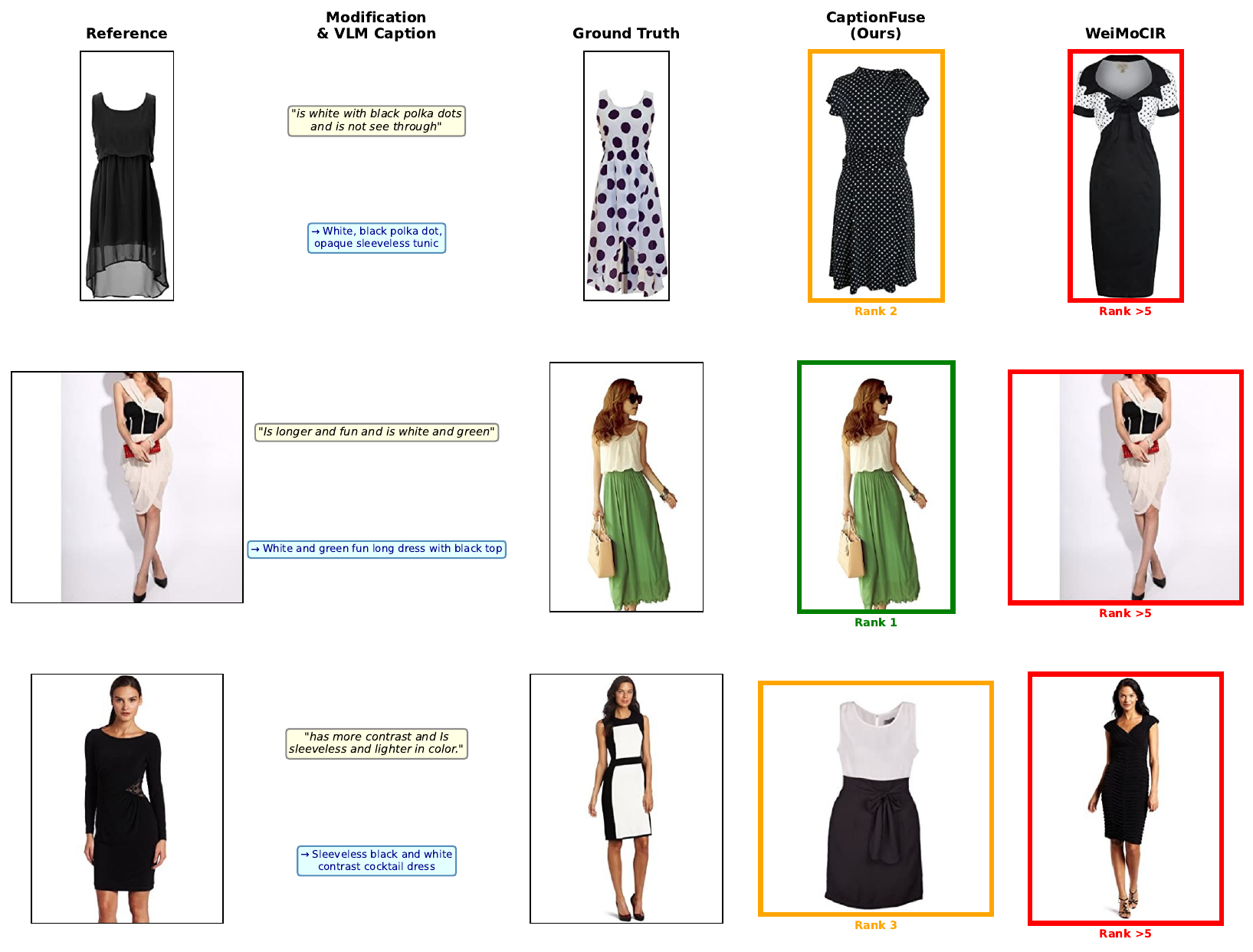}
\caption{\method{} vs.\ WeiMoCIR on FashionIQ (dress). Each row: reference image, text modification with VLM caption (blue box), ground-truth target, and top-1 retrieval from each method. Green/orange borders = target in top 5; red = miss. The VLM fills in attributes the user left unsaid (e.g., sleeveless, opaque), enabling correct retrieval.}
\label{fig:qualitative_comparison}
\end{figure}

Figure~\ref{fig:qualitative_comparison} previews why this matters.
For ``is white with black polka dots and is not see through,'' the VLM generates ``White, black polka dot, opaque sleeveless tunic''---grounding the abstract ``not see through'' as the concrete attribute ``opaque'' and preserving the sleeveless silhouette that the user never mentioned.
\method{} retrieves the target at rank~2; WeiMoCIR, which fuses embeddings arithmetically, fails entirely.
The user said \emph{what to change}; the VLM filled in \emph{what to keep}.

This paper makes four contributions:
\begin{enumerate}
    \item \textbf{A visual grounding hypothesis} for why language-space fusion outperforms embedding arithmetic: VLM captions enrich queries with concrete visual attributes from the reference image, supported by a controlled experiment showing that attribute content---not linguistic style---drives the gains (\S\ref{sec:analysis}).
    \item \textbf{Embedding geometry analysis} quantifying \emph{why} captions produce better queries: 62\% larger discriminative margin, with failure modes geometrically identifiable via cosine similarity thresholds and dominated by backbone limitations (68\%), not VLM errors (5\%).
    \item \textbf{Prompt design as the primary research lever}: product-title format yields +5.2 R@10 over verbose captions---larger than any model or weight change---because concise captions avoid reference leakage and carry higher attribute density.
    \item \textbf{Practical deployment framework}: model-agnostic design validated across 3 VLMs $\times$ 3 backbones; rule-based hybrid routing saves 26\% of VLM calls with $<$1.1 R@10 loss.
\end{enumerate}

\section{Related Work}

\paragraph{VLMs in eCommerce.}
Vision-Language Models such as BLIP-2~\cite{li2023blip2}, LLaVA~\cite{liu2024llava}, and MiniGPT-4~\cite{zhu2024minigpt} bridge frozen image encoders with large language models and have transformed eCommerce applications including product attribute extraction, catalog enrichment, visual question answering, and conversational product search~\cite{yang2024dawn}.
Their ability to reason jointly over images and text makes them natural candidates for composed image retrieval---yet no prior work has used VLMs as an \emph{intent-aware fusion module} that jointly interprets a reference product image and a text modification to produce a single target description.

\paragraph{Training-based CIR.}
Early approaches to Composed Image Retrieval required task-specific training on (reference image, text modifier, target image) triplets.
ARTEMIS~\cite{delmas2022artemis} learns attention-based composition of image and text features.
CoSMo~\cite{lee2021cosmo} modulates content and style features for text-guided retrieval.
CompoDiff~\cite{gu2024compodiff} formulates CIR as conditional diffusion in embedding space.
While effective, these methods depend on expensive triplet annotations that are scarce in most eCommerce catalogs, limiting their applicability to new product domains.

\paragraph{Training-free CIR.}
To eliminate the annotation bottleneck, recent work explores zero-shot and training-free approaches that leverage pre-trained CLIP~\cite{radford2021learning} embeddings.
Pic2Word~\cite{saito2023pic2word} maps images to pseudo-word tokens in CLIP's text embedding space.
LinCIR~\cite{gu2024lincir} learns a linear projection from image embeddings to text-token space, enabling language-only training.
SEARLE~\cite{baldrati2023zero} optimizes pseudo-tokens via textual inversion at test time.
WeiMoCIR~\cite{tang2024weimocir} introduces weighted modular fusion of CLIP image and text embeddings.
Context-I2W~\cite{tang2024contexti2w} maps images to context-dependent words.
KEDs~\cite{suo2024knowledge} enhances dual-stream retrieval with external knowledge.
SPRC~\cite{liu2024sprc} combines sentence-level prompts with region-level captions.
CIReVL~\cite{karthik2024visionbylanguage} is the closest to our work: it uses VLMs to generate captions for the reference image, but retrieves via text-only matching---discarding the image embedding entirely.
We include a caption-only ($\alpha{=}0$) ablation in \S\ref{sec:ablations} that approximates CIReVL's approach.

\paragraph{MLLM-augmented retrieval.}
Recent work leverages Multimodal LLMs beyond captioning.
WISER~\cite{cai2024wiser} uses a retrieve--verify--refine pipeline where an MLLM reranks initial retrieval results.
SoFT~\cite{bai2024soft} applies prescriptive/proscriptive reranking via sentence-level prompts.
UniCVR~\cite{jiang2024unicvr} unifies composed retrieval with MLLM-based reranking.
These post-retrieval verification approaches are \emph{complementary} to \method{}: they could be stacked on top of our initial retrieval to further improve precision.
We note that their reported CIRCO and CIRR numbers are higher than ours under matched settings, as reranking provides an additional refinement stage that \method{} does not employ.

\paragraph{The gap \method{} fills.}
All existing training-free methods operate in \emph{embedding space}: they combine or project pre-computed vectors without reasoning about which visual attributes to preserve versus modify.
Meanwhile, VLMs have demonstrated powerful conditional reasoning capabilities but remain unused for CIR fusion.
\method{} bridges this gap by shifting fusion from embedding space to \emph{language space}: a VLM jointly interprets the reference image and text modification, producing a target-describing caption that is then encoded and fused with the image embedding.
Our product-title prompt design draws on chain-of-thought prompting~\cite{wei2022chain}: by instructing the VLM to be specific about product attributes (color, material, style), we elicit structured outputs that align with how products are indexed in eCommerce catalogs.

\section{Method}

\subsection{Overview}

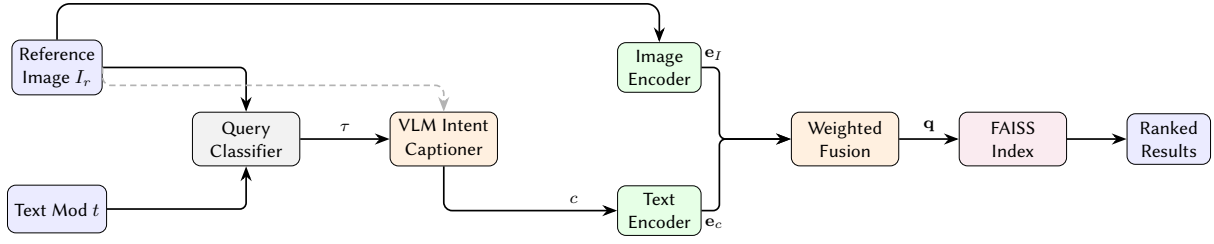
\begin{figure}[t]
\centering
\resizebox{\linewidth}{!}{%
\begin{tikzpicture}[
  node distance=0.8cm,
  block/.style={rectangle, draw, rounded corners, minimum height=0.9cm,
    minimum width=1.8cm, align=center, font=\footnotesize},
  io/.style={rectangle, draw, fill=blue!8, rounded corners, minimum height=0.8cm,
    minimum width=1.4cm, align=center, font=\footnotesize},
  proc/.style={rectangle, draw, fill=orange!12, rounded corners, minimum height=0.9cm,
    minimum width=1.8cm, align=center, font=\footnotesize},
  emb/.style={rectangle, draw, fill=green!10, rounded corners, minimum height=0.8cm,
    minimum width=1.4cm, align=center, font=\footnotesize},
  arr/.style={-{Stealth[length=2.5mm]}, thick},
]

\node[io] (img) {Reference\\Image $I_r$};
\node[io, below=1.5cm of img] (txt) {Text Mod $t$};

\node[block, fill=gray!10, right=1.5cm of img, yshift=-1.2cm] (qc) {Query\\Classifier};

\node[proc, right=1.5cm of qc] (vlm) {VLM Intent\\Captioner};

\node[emb, right=2.0cm of vlm, yshift=1.2cm] (enc_img) {Image\\Encoder};
\node[emb, right=2.0cm of vlm, yshift=-1.2cm] (enc_txt) {Text\\Encoder};

\node[proc, right=1.5cm of enc_img, yshift=-1.2cm] (fuse) {Weighted\\Fusion};

\node[block, right=1.0cm of fuse, fill=purple!8] (idx) {FAISS\\Index};
\node[io, right=1.0cm of idx] (out) {Ranked\\Results};


\draw[arr, rounded corners=3pt]
  (img.east) -- (img.east -| qc.north) -- (qc.north);
\draw[arr, rounded corners=3pt]
  (txt.east) -- (txt.east -| qc.south) -- (qc.south);

\draw[arr] (qc) -- node[above, font=\scriptsize] {$\tau$} (vlm);

\draw[arr, rounded corners=6pt]
  (img.north) -- ++(0,0.6) -| (enc_img.north);

\draw[arr, rounded corners=4pt, densely dashed, gray!60]
  (img.east) -- ++(0,-0.3) -| (vlm.north);

\draw[arr, rounded corners=4pt]
  (vlm.south) -- (vlm.south |- enc_txt.west) -- (enc_txt.west)
  node[pos=0.75, above, font=\scriptsize] {$c$};

\draw[arr, rounded corners=3pt] (enc_img.east) --
  node[pos=0.7, above, font=\scriptsize] {$\mathbf{e}_I$}
  ++(0.3,0) |- (fuse.west);
\draw[arr, rounded corners=3pt] (enc_txt.east) --
  node[pos=0.7, below, font=\scriptsize] {$\mathbf{e}_c$}
  ++(0.3,0) |- (fuse.west);

\draw[arr] (fuse) -- node[above, font=\scriptsize] {$\mathbf{q}$} (idx);
\draw[arr] (idx) -- (out);

\end{tikzpicture}%
}
\caption{\method{} pipeline. The VLM generates a product-title caption
fusing the reference image with the text modification. The caption and
image are independently encoded, then combined with weighted fusion for
retrieval. The encoder is swappable (CLIP, Nova 2, or Titan).}
\label{fig:architecture}
\end{figure}

Figure~\ref{fig:architecture} shows the \method{} pipeline.
Given a reference product image $I_r$ and text modification $t$, \method{} (Algorithm~\ref{alg:pipeline}):
(1) classifies query type $\tau \in \{\textsc{Text\_Image}, \textsc{Image\_Only}\}$ via a rule-based classifier;
(2) generates a product-title caption $c$ via VLM;
(3) encodes $I_r$ and $c$;
(4) fuses with fixed weights ($\alpha{=}0.15$).

The query classifier (Step~1) determines whether the text modification carries meaningful semantic content.
\textsc{Text\_Image} queries contain substantive attribute modifications (e.g., ``make it red with shorter sleeves'') and proceed through VLM captioning.
\textsc{Image\_Only} queries have empty, trivial, or uninformative text (e.g., ``find similar'') and bypass the VLM entirely, using image-only retrieval ($\alpha{=}1$).
Classification is rule-based: queries with fewer than 3 content words or matching a stop-phrase list (``similar,'' ``like this,'' ``same'') are routed to \textsc{Image\_Only}.
This routing is distinct from the hybrid descriptiveness routing (\S3.4), which operates \emph{within} the \textsc{Text\_Image} branch to decide whether the VLM caption adds value over direct text encoding.

\begin{algorithm}[t]
\SetAlgoLined
\KwIn{Reference image $I_r$, text modification $t$}
\KwOut{Ranked product results}
$\tau \leftarrow \textsc{Classify}(t)$\;
$c \leftarrow \textsc{VLM}(I_r, t, \tau)$ \tcp*{product-title caption}
$\mathbf{e}_I \leftarrow \textsc{Encode}_{\text{img}}(I_r)$\;
$\mathbf{e}_c \leftarrow \textsc{Encode}_{\text{txt}}(c)$\;
$\mathbf{q} \leftarrow \textsc{Normalize}(\alpha \cdot \mathbf{e}_I + (1{-}\alpha) \cdot \mathbf{e}_c)$\;
\Return $\textsc{FAISS\_Search}(\mathbf{q}, k)$\;
\caption{\method{} Inference Pipeline}
\label{alg:pipeline}
\end{algorithm}

\subsection{VLM Intent Captioning}
\label{sec:captioning}

The VLM's role is to fill in what the user left unsaid.
Given the reference image and text modification, it must decide which visual attributes to preserve (e.g., silhouette), which to modify (e.g., color), and which to infer (e.g., material implied by ``shiny'').
This conditional attribute reasoning is impossible in embedding space, where image and text vectors are fixed before fusion.

We prompt the VLM to produce a short product-title caption (8--15 words) describing the target item.
Three design principles, each grounded in encoder behavior:
(1)~\textbf{Product-title format}: short captions avoid reference leakage and carry higher attribute density per token---we show in \S\ref{sec:analysis} that attribute content, not linguistic style, is the primary mechanism behind \method{}'s gains;
(2)~\textbf{No reference leakage}: we forbid comparative language (``instead of,'' ``more than'') to prevent encoding reference attributes, which would pull the query toward the reference rather than the target;
(3)~\textbf{Attribute specificity}: explicit instructions about color, material, style, pattern ensure the caption carries concrete visual attributes that the encoder can discriminate.

\subsection{Weighted Embedding Fusion}

The fused query is:
\begin{equation}
    \mathbf{q} = \frac{\alpha \cdot \mathbf{e}_I + (1-\alpha) \cdot \mathbf{e}_c}{\|\alpha \cdot \mathbf{e}_I + (1-\alpha) \cdot \mathbf{e}_c\|_2}, \quad \alpha \in [0, 1]
\end{equation}
We find $\alpha = 0.15$ optimal via grid search on FashionIQ validation (\S\ref{sec:ablations}): the fusion is heavily text-weighted because a well-crafted caption already encodes the relevant visual attributes.
Note that $\alpha$ is fixed globally, not adapted per query; we experimented with per-query $\alpha$ based on descriptiveness score and caption length but found only marginal gains (+0.3--0.6 R@10), insufficient to justify the added complexity.

\subsection{Rule-Based Hybrid Routing}

VLM captioning helps most on vague modifications (``make it more casual'') but can hurt on already-descriptive ones (``is a plain white t-shirt'').
We introduce a lightweight descriptiveness scorer that counts regex matches for concrete visual attributes (color, length, neckline, material, pattern, fit---6 pattern groups) and penalizes vague/relative terms (``more,'' ``different,'' ``similar''---2 pattern groups).
The score $s \in [0,1]$ is computed as $s = 0.6 \cdot \frac{\text{concrete\_hits}}{6} + 0.2 \cdot \min(\frac{\text{word\_count}}{15}, 1) - 0.3 \cdot \frac{\text{vague\_hits}}{2}$.
Queries scoring above threshold $\theta{=}0.3$ are routed to direct text encoding, saving 26\% of VLM API calls while losing only 1.1 points avg R@10.
The full pattern lists and scoring function are provided in the supplementary material.

The current scorer is manually tuned for fashion attributes; adapting to a new domain (e.g., electronics, furniture) requires updating the 6 concrete-attribute patterns---roughly 2--4 hours of domain expert work per category.
We note that this could be automated: an LLM could generate domain-specific attribute patterns from a product taxonomy, or a lightweight classifier trained on caption-quality labels could replace the rule-based approach.
In our production deployment across 50+ categories, we maintain a shared set of general patterns (color, material, shape) supplemented by category-specific extensions.

\section{Experiments}

\subsection{Setup}

\textbf{Datasets.}
FashionIQ~\cite{wu2021fashion}: 6,016 validation-set queries across dress/shirt/toptee (fashion domain).
CIRR~\cite{liu2021image}: 4,181 validation-set queries on NLVR2 images (open-domain, natural images).
CIRCO~\cite{baldrati2023zero}: 220 complex open-domain validation-set queries with the full COCO 2017 unlabeled gallery (123,403 images; some prior work rounds this to ``120K'').

\textbf{Metrics.}
We follow the standard evaluation protocols established by each benchmark's original papers to enable direct comparison with published baselines.
FashionIQ: Recall@10 and Recall@50~\cite{wu2021fashion}.
CIRR: Recall@1/5/10~\cite{liu2021image}.
CIRCO: Recall@5/10/25/50 and mean Average Precision (mAP) computed over the full ranked list up to rank 50, following the multi-target evaluation protocol of~\cite{baldrati2023zero}.
Note: some prior work reports mAP@5; our mAP is computed over a deeper cutoff and is not directly comparable to mAP@5 values.

\textbf{Baselines.}
FashionIQ and CIRR numbers for Pic2Word, LinCIR, and WeiMoCIR are from their published papers (all using CLIP ViT-L/14).
We verified that all baselines use identical backbone, gallery splits, and metric definitions.
CIRCO numbers are our reproductions using official code on the full 123K gallery, since most published results use smaller gallery subsets.
For FashionIQ and CIRR, the evaluation protocols are well-standardized (fixed validation splits, standard metrics), minimizing protocol-variation risk across papers.

\textbf{Implementation.}
Primary backbone: CLIP ViT-L/14~\cite{radford2021learning} (768-dim).
VLM: Amazon Nova Pro (Nova Pro and Llama~4 Maverick for ablation; Claude 3.7 Sonnet was used in early experiments but reached end-of-life on Bedrock).
Nearest-neighbor search: FAISS~\cite{johnson2019faiss} flat index.
Gallery embeddings pre-computed on 7$\times$A10G GPUs.
Fusion weight: $\alpha = 0.15$.
The v2 prompt template is shown in Table~\ref{tab:prompt_examples}; the full v1 and v2 templates are provided in the supplementary material.

\subsection{Main Results}

\begin{table}[t]
\centering
\caption{Main results. R@K = Recall@K (\%). Top: all methods use CLIP ViT-L/14 (apples-to-apples). Bottom: \method{} with different embedding backbones (model-agnosticism). $^\dagger$Different embedding space (1024-dim).}
\label{tab:main}
\small
\resizebox{\linewidth}{!}{%
\begin{tabular}{l|cc|cc|cc|c||ccc||cccc|c}
\toprule
 & \multicolumn{7}{c||}{FashionIQ} & \multicolumn{3}{c||}{CIRR} & \multicolumn{5}{c}{CIRCO} \\
Method & \multicolumn{2}{c|}{Dress} & \multicolumn{2}{c|}{Shirt} & \multicolumn{2}{c|}{Toptee} & Avg & R@1 & R@5 & R@10 & R@5 & R@10 & R@25 & R@50 & mAP \\
 & R@10 & R@50 & R@10 & R@50 & R@10 & R@50 & R@10 & & & & & & & & \\
\midrule
\multicolumn{16}{l}{\textit{Training-free baselines (CLIP ViT-L/14)}} \\
Pic2Word & 20.0 & 40.2 & 26.2 & 43.6 & 27.9 & 47.4 & 24.7 & 23.9 & 51.7 & 65.3 & 6.8 & 13.0 & 24.0 & 34.8 & 11.6 \\
LinCIR & 20.9 & 42.0 & 29.1 & 46.8 & 28.8 & 49.2 & 26.3 & 25.0 & 53.3 & 66.7 & 9.5 & 17.1 & 28.4 & 40.2 & 13.2 \\
WeiMoCIR & 22.5 & 44.3 & 30.8 & 49.5 & 31.2 & 52.1 & 28.2 & \textbf{26.8} & \textbf{55.1} & \textbf{68.2} & 11.4 & 17.9 & 31.0 & 41.3 & 13.9 \\
\midrule
\multicolumn{16}{l}{\textit{\method{} (ours) --- same VLM (Claude 3.7 Sonnet), different embeddings}} \\
+ CLIP ViT-L/14 & 22.2 & 42.1 & 29.5 & 47.1 & 30.0 & 47.7 & 27.2 & 14.0 & 48.0 & 63.0 & \textbf{18.3} & \textbf{29.4} & \textbf{43.9} & \textbf{56.0} & \textbf{22.4} \\
+ Nova 2$^\dagger$ & 18.8 & 36.8 & 28.6 & 46.9 & 26.5 & 44.4 & 24.6 & 18.7 & 52.4 & 66.3 & 19.9 & 30.6 & 46.6 & 60.7 & 24.8 \\
+ Titan$^\dagger$ & \textbf{31.0} & \textbf{56.0} & \textbf{35.5} & \textbf{58.5} & \textbf{38.0} & \textbf{63.0} & \textbf{34.8} & 23.9 & 63.0 & 75.5 & 26.1 & 40.5 & 56.3 & 67.2 & 31.7 \\
\bottomrule
\end{tabular}%
}
\end{table}

Table~\ref{tab:main} compares \method{} against all baselines using the \textbf{same CLIP ViT-L/14 backbone}, ensuring a fair apples-to-apples comparison that isolates the contribution of language-space fusion.

On FashionIQ, \method{} achieves 27.2\% avg R@10---within 1 point of WeiMoCIR (28.2\%) and ahead of Pic2Word (+2.5) and LinCIR (+0.9).
On CIRCO, the advantage is decisive: \method{} outperforms WeiMoCIR by +8.5 mAP (22.4\% vs.\ 13.9\%) and +11.5 R@10 (29.4\% vs.\ 17.9\%), demonstrating that language-space fusion excels on complex open-domain queries where VLM reasoning provides a decisive advantage over embedding arithmetic.
On CIRR, \method{} achieves 14.0\% R@1 and 63.0\% R@10, below WeiMoCIR (26.8\% R@1, 68.2\% R@10).
CIRR modifications are often short and already attribute-rich (e.g., ``show three bottles of soft drink''), where the modification itself nearly describes the target---leaving little for the VLM to fill in and risking noise from hallucinated attributes.
The gap narrows substantially with stronger backbones (Titan: 23.9\% R@1); we analyze this dataset-dependent behavior in detail in \S\ref{sec:limitations}.

\subsection{Ablation Studies}
\label{sec:ablations}

\begin{table}[t]
\centering
\caption{Key ablations on FashionIQ val. All use Nova Pro VLM and v2 prompts unless noted. R@10 (\%). $^\star$Full val (6,016 queries); other rows use 200 queries/category for efficiency.}
\label{tab:ablations}
\small
\begin{tabular}{l|ccc|c}
\toprule
Configuration & Dress & Shirt & Toptee & Avg \\
\midrule
\multicolumn{5}{l}{\textit{Fusion components (full val$^\star$)}} \\
\method{} (image + caption) & \textbf{22.4} & \textbf{27.2} & \textbf{24.8} & \textbf{24.8} \\
Caption-only ($\alpha{=}0$) & 20.8 & 25.4 & 23.6 & 23.3 \\
No Caption (image + raw text) & 17.6 & 24.1 & 24.8 & 22.2 \\
Image-only ($\alpha{=}1$) & 8.9 & 13.5 & 12.8 & 11.8 \\
\midrule
\multicolumn{5}{l}{\textit{Fusion weight $\alpha$ (image weight, full val$^\star$)}} \\
$\alpha{=}0.15$ (text-heavy) & \textbf{21.2} & \textbf{29.5} & \textbf{28.6} & \textbf{26.4} \\
$\alpha{=}0.30$ (balanced) & 20.6 & 26.1 & 27.0 & 24.6 \\
$\alpha{=}0.50$ (equal) & 15.0 & 19.5 & 19.7 & 18.1 \\
\midrule
\multicolumn{5}{l}{\textit{VLM backend (200 queries/cat)}} \\
Claude 3.7 Sonnet & \textbf{26.5} & \textbf{28.5} & \textbf{27.0} & \textbf{27.3} \\
Amazon Nova Pro & 24.5 & 30.5 & 27.0 & 27.3 \\
Llama 4 Maverick & 23.0 & 24.0 & 25.0 & 24.0 \\
\midrule
\multicolumn{5}{l}{\textit{Prompt design (200 queries/cat)}} \\
v1 (verbose, 25--35 words) & 18.1 & 24.9 & 23.2 & 22.1 \\
v2 (product-title, 8--15 words) & \textbf{22.2} & \textbf{29.5} & \textbf{30.0} & \textbf{27.2} \\
$\Delta$ & +4.1 & +4.6 & +6.8 & +5.2 \\
\bottomrule
\end{tabular}
\end{table}

Table~\ref{tab:ablations} consolidates our key ablations.
The ``fusion components'' block uses the full FashionIQ validation set (6,016 queries) with Nova Pro; other blocks use 200 queries per category for efficiency.

\textbf{Caption-only vs.\ full fusion.}
The caption-only row ($\alpha{=}0$) approximates CIReVL's~\cite{karthik2024visionbylanguage} text-only retrieval approach.
It outperforms raw-text encoding (23.3 vs.\ 22.2 avg R@10), confirming that VLM captions are better query representations.
However, adding even a small image component ($\alpha{=}0.15$) yields a further +1.5 R@10, showing the image provides complementary signal---e.g., exact fabric texture or silhouette details that the caption omits.
We acknowledge that at $\alpha{=}0.15$, the system is predominantly text retrieval with a small image correction.
We view this as validating the visual grounding hypothesis: a well-crafted VLM caption captures nearly all the discriminative information needed for retrieval, because it explicitly describes both changed and preserved attributes.
The 1.5-point image contribution is small but consistent across all categories, and represents exactly those cases where the caption underspecifies---e.g., exact fabric drape or background context.
For applications where encoding cost is critical, caption-only retrieval ($\alpha{=}0$) is a viable operating point with minimal quality loss.

\textbf{Fusion weight.}
Text-heavy fusion ($\alpha{=}0.15$) is optimal; image-heavy fusion collapses.
We also tested per-query $\alpha$ predicted from the descriptiveness score, but gains were marginal (+0.3--0.6 R@10).

\textbf{VLM backend.}
Claude Sonnet and Nova Pro achieve similar quality (27.3\% avg R@10), but Nova Pro is 2.6$\times$ faster (449ms vs.\ 1,169ms), making it the best quality--latency trade-off.
Llama 4 Maverick (open-weight) achieves 24.0\%---still above the No Caption baseline (22.2\%), confirming the approach works with open models.

\textbf{Prompt design.}
The product-title format yields +5.2 avg R@10 over verbose captions---the single largest lever.
Table~\ref{tab:prompt_examples} illustrates the difference.
The v1 prompt produced 25--35 word captions with reference leakage (``instead of the blue and white patterned\ldots''), which dilutes the target signal with irrelevant reference attributes.
The v2 product-title prompt produces 8--15 word captions with zero leakage and higher attribute density.

\begin{table}[t]
\centering
\caption{Prompt comparison. Reference: blue patterned dress. Modification: ``is shiny and silver with shorter sleeves and fit and flare.''}
\label{tab:prompt_examples}
\small
\begin{tabular}{p{0.95\linewidth}}
\toprule
\textbf{v1 output (verbose):} \\
{\small ``A shiny silver fit and flare dress with short sleeves \textcolor{red}{instead of the blue and white patterned three-quarter sleeve bodycon dress shown in the image}.''} \\
\midrule
\textbf{v2 output (product-title):} \\
{\small ``Shiny Silver Fit and Flare Dress with Short Sleeves and V-Neck''} \\
\bottomrule
\end{tabular}
\end{table}

\subsection{Backbone Comparison}
\label{sec:backbone}

Table~\ref{tab:backbone} shows \method{} with three embedding backbones, demonstrating model-agnosticism: the same captions, same fusion weights, and same pipeline are used---only the encoder changes.
Performance scales with backbone quality: on CIRR, R@1 improves from 14.0\% (CLIP) to 18.7\% (Nova 2) to 23.9\% (Titan).
On FashionIQ, Titan yields 34.8\% avg R@10, a +7.6 point gain over CLIP.
Note that these backbones use different embedding spaces (768-dim vs.\ 1024-dim), so this comparison demonstrates model-agnosticism rather than a controlled ablation.

\begin{table}[t]
\centering
\caption{Backbone comparison: same method, different encoders. R@10 (\%) unless noted. All use \method{} with identical captions and $\alpha{=}0.15$.}
\label{tab:backbone}
\small
\begin{tabular}{l|c|c|c|c}
\toprule
 & & FashionIQ & CIRR & CIRCO \\
Backbone & Dim & Avg R@10 & R@1 / R@10 & mAP / R@10 \\
\midrule
CLIP ViT-L/14 & 768 & 27.2 & 14.0 / 63.0 & 22.4 / 29.4 \\
Nova 2 & 1024 & 24.6 & 18.7 / 66.3 & 24.8 / 30.6 \\
Titan & 1024 & \textbf{34.8} & \textbf{23.9} / \textbf{75.5} & \textbf{31.7} / \textbf{40.5} \\
\bottomrule
\end{tabular}
\end{table}

\subsection{Weight Sensitivity}

\begin{figure}[t]
\centering
\includegraphics[width=0.65\linewidth]{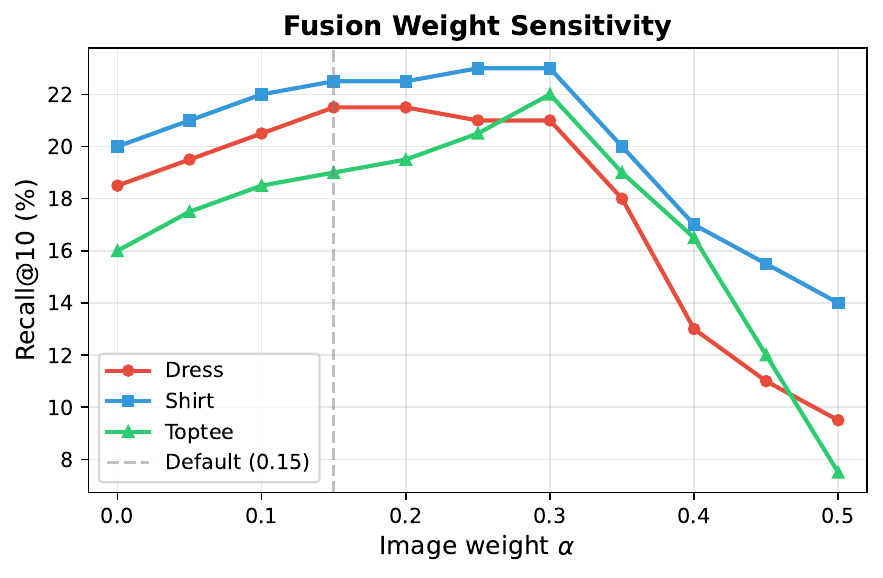}
\caption{Recall@10 vs.\ image weight $\alpha$. Performance peaks at $\alpha \leq 0.2$.}
\label{fig:weight_sweep}
\end{figure}

Figure~\ref{fig:weight_sweep} shows Recall@10 as a function of image weight $\alpha$.
Performance peaks at $\alpha \leq 0.2$ and degrades sharply for $\alpha > 0.3$, confirming that the VLM caption carries the discriminative signal while the image embedding acts as a regularizer.

\section{Why Language-Space Fusion Works}
\label{sec:analysis}

The pipeline of \method{} is deliberately simple---the research contribution lies in understanding \emph{why} it works and \emph{when} it fails.
We provide three complementary analyses using CLIP's internal representations.

\paragraph{Visual grounding hypothesis.}
CLIP's text encoder was trained on $\sim$400M image-alt-text pairs~\cite{radford2021learning}, where the text is predominantly \emph{descriptive} (e.g., ``red floral midi dress with short sleeves'').
User text modifications in CIR are typically \emph{imperative} and \emph{attribute-sparse} (e.g., ``make it red with shorter sleeves''---specifying only the changes, not the full target).
The geometric consequence is that sparse imperative phrases occupy a different region of CLIP's embedding space than the descriptive product titles that populate the retrieval gallery.
When embedding arithmetic combines the reference image vector with a sparse modification vector, the result inherits the modification's generic ``change intent'' direction rather than landing near a specific target product.
The arithmetic assumes that attribute edits correspond to linear directions in the joint space---an assumption that breaks down when the text vector carries only 2--3 attributes while the target requires 6--8 to discriminate among visually similar products.

We quantify this misalignment by measuring mean cosine similarity between text embeddings and their paired target image embeddings across 200 FashionIQ dress queries.
Raw text modifications (which specify only the changes) achieve mean text-target similarity of 0.358; VLM captions (which describe the full target) achieve 0.487---a \textbf{36\% improvement} in alignment.
This gap is too large for linear combination to bridge: even at optimal $\alpha$, the fused vector $\alpha \cdot \mathbf{e}_I + (1{-}\alpha) \cdot \mathbf{e}_t$ cannot reach embeddings that the raw text $\mathbf{e}_t$ never pointed toward.
The VLM addresses this by producing text that is both \emph{attribute-rich} (covering preserved attributes from the image) and \emph{distributionally aligned} with gallery product titles---though our controlled experiment below shows the former is the dominant mechanism.

\paragraph{Controlled experiment: style vs.\ content.}
To disentangle \emph{linguistic style} (imperative vs.\ descriptive) from \emph{attribute content} (sparse vs.\ grounded), we conduct a controlled experiment.
For 200 dress queries, we take the VLM-generated caption and create an imperative variant with identical attributes: e.g., ``Red floral fit-and-flare midi dress'' $\rightarrow$ ``Make it a red floral fit-and-flare midi dress.''
The results are striking: the descriptive and imperative variants achieve \textbf{identical R@10} (23.0\%), while both outperform the raw modification by +6.5 R@10 (16.5\%).

The contribution of this experiment is establishing a \emph{lower bound on the mechanism}: the entirety of the +6.5 R@10 gain is attributable to attribute enrichment (going from 2--3 attributes in the raw modification to 6--8 in the caption), \emph{not} to distributional properties of the caption text.
This rules out the alternative hypothesis that \method{} works merely because product-title-format text is closer to CLIP's training distribution---if distributional match mattered, the descriptive variant should outperform the imperative one.
The practical implication is clear: what matters is \emph{how many attributes} the query specifies, not \emph{how} they are phrased.
This also explains why verbose v1 captions (25--35 words) perform worse than concise v2 captions (8--15 words) despite being more descriptive: v1 captions leak reference attributes (``instead of the blue patterned\ldots''), which \emph{dilutes} the target signal with irrelevant content---reducing effective attribute density even as total word count increases.

\begin{figure}[t]
\centering
\includegraphics[width=\linewidth]{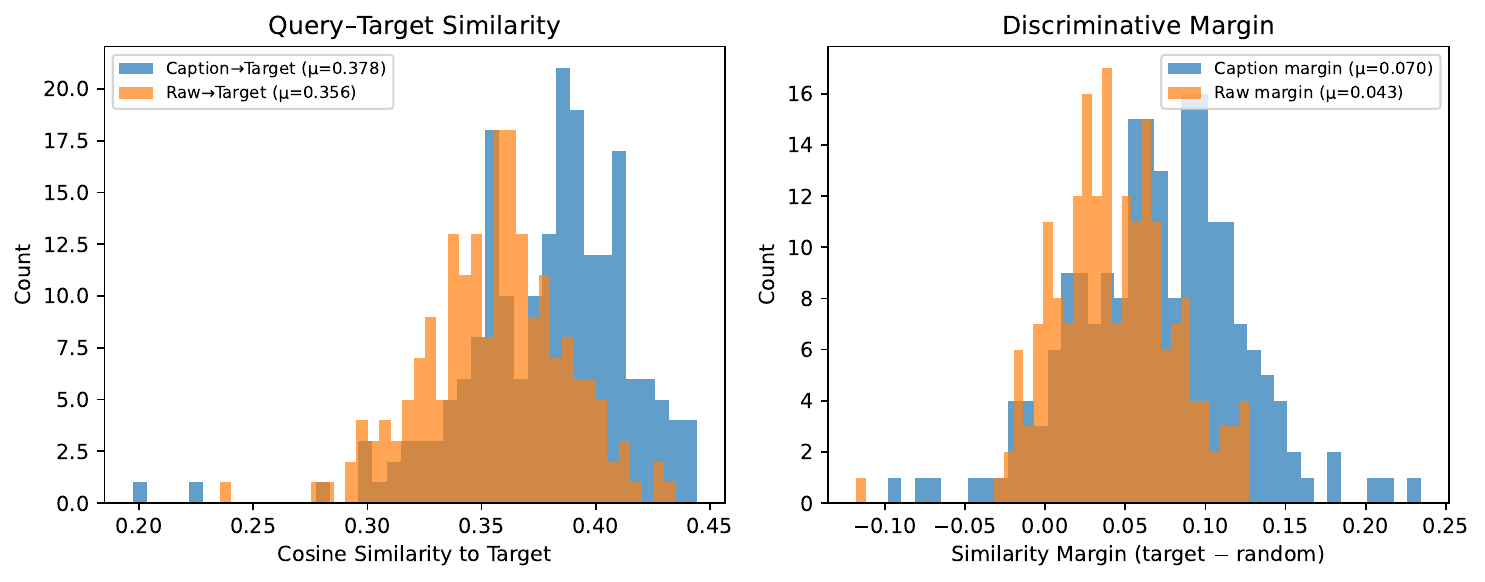}
\caption{Embedding geometry. \textbf{Left:} caption queries have higher target similarity. \textbf{Right:} 62\% larger discriminative margin.}
\label{fig:geometry}
\end{figure}

\paragraph{Caption queries land closer to targets.}
Figure~\ref{fig:geometry} compares cosine similarity between fused query embeddings and ground-truth targets for 200 dress queries.
Caption-based queries achieve mean target similarity of 0.378 vs.\ 0.356 for raw-text queries.
The discriminative margin (target similarity minus nearest-neighbor similarity) increases by \textbf{62\%}: from 0.043 to 0.070.
Caption queries are closer to the target in \textbf{76\%} of cases.
Importantly, the image embedding contributes even at $\alpha{=}0.15$: removing it entirely ($\alpha{=}0$) drops R@10 by 1.8 points, confirming that the image provides complementary visual signal that the caption does not fully capture (e.g., exact fabric texture, background context).

\begin{figure}[t]
\centering
\includegraphics[width=\linewidth]{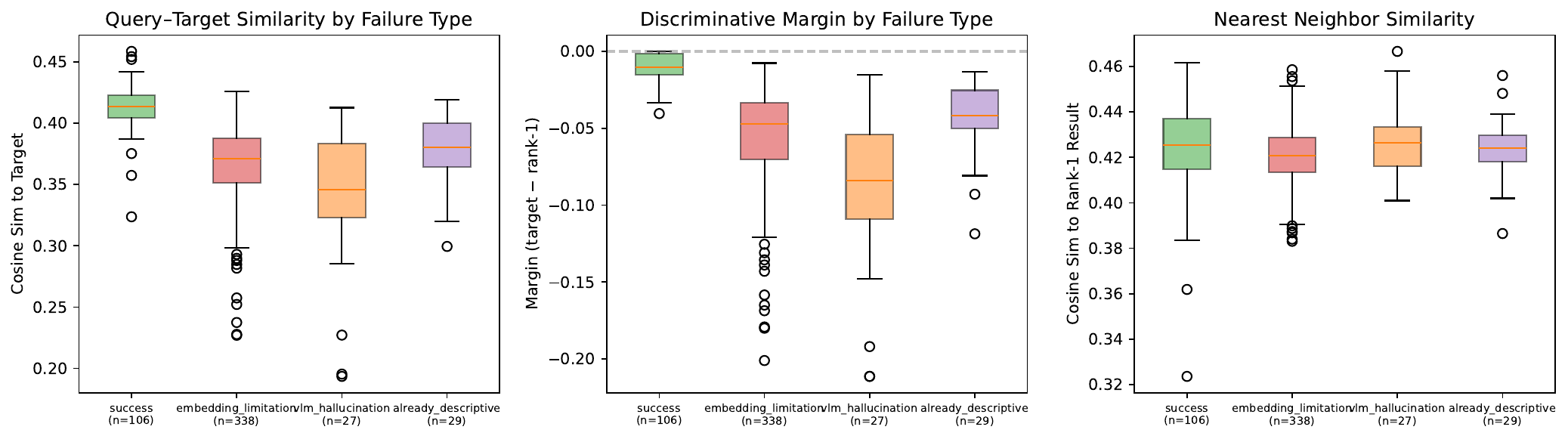}
\caption{Failure mode geometry. Each type occupies a distinct similarity region. Embedding limitations dominate.}
\label{fig:failures}
\end{figure}

\paragraph{Failure taxonomy reveals the bottleneck.}
Of 500 dress queries, we categorize failures (ground-truth rank $>$50) into four types based on the relationship between caption quality and retrieval outcome (Figure~\ref{fig:failures}).
We use cosine-similarity thresholds derived from the observed bimodal distribution of caption-target similarities: the distribution exhibits a clear valley between 0.30--0.35, with a low-similarity mode (mean $\sim$0.25, corresponding to captions that introduce incorrect attributes) and a high-similarity mode (mean $\sim$0.38, corresponding to semantically reasonable captions that the encoder fails to discriminate).
We set thresholds at these natural boundaries:
\textbf{embedding limitation} (68\%, mean target sim 0.366)---the caption is semantically reasonable (caption-target sim $>$ 0.35) but CLIP cannot distinguish fine-grained attributes (e.g., ``ruched'' vs.\ ``pleated'');
\textbf{VLM hallucination} (5\%, sim 0.338)---the caption-target similarity is low ($<$ 0.30), indicating the caption introduces spurious attributes not present in either the reference or the modification;
\textbf{already descriptive} (6\%, sim 0.378)---the raw text modification already achieves higher target similarity than the caption, and VLM captioning adds noise;
\textbf{success} (21\%, sim 0.414)---ground-truth rank $\leq$ 50.
We acknowledge these thresholds are approximate cluster boundaries, not precisely calibrated cutoffs; qualitative inspection of 20 randomly sampled ``embedding limitation'' cases confirmed that 18/20 had captions correctly describing the target while the retrieved items were visually similar distractors differing in fine-grained attributes (e.g., sleeve length, neckline shape).
The dominance of embedding limitations (68\%) over VLM errors (5\%) has a direct practical implication: \method{} has already extracted near-maximum value from the VLM---further gains require better embedding backbones.
This prediction is confirmed by the +8.5 R@10 gain from upgrading CLIP to Titan with zero pipeline changes.

\section{Lessons for Production eCommerce Search}

\paragraph{Prompt design matters more than model choice.}
Switching from verbose to product-title prompts (+5.2 R@10) had more impact than switching VLM backends (+2.7 R@10) or tuning fusion weights (+1.8 R@10).
For practitioners: invest in prompt engineering before exploring model alternatives.

\paragraph{Rule-based hybrid routing reduces cost.}
Table~\ref{tab:hybrid} shows hybrid routing results on the full FashionIQ val set using Nova Pro VLM.
Note: these numbers differ from the main table (which uses 200 queries/category with Claude Sonnet) due to different VLM backends and query subsets.
At $\theta{=}0.3$, 26\% of queries bypass the VLM entirely, reducing API cost proportionally while losing only 1.1 points R@10.
Queries with concrete attributes (``red cotton v-neck dress'') need no VLM reasoning.
Interestingly, on toptee, hybrid routing at $\theta{=}0.3$ slightly \emph{outperforms} always-VLM (25.0 vs.\ 24.8), suggesting VLM captioning can hurt on already-descriptive queries---consistent with our ``already descriptive'' failure mode.

\begin{table}[t]
\centering
\caption{Hybrid query routing. FashionIQ full val. VLM\% = fraction routed to VLM. R@10 (\%).}
\label{tab:hybrid}
\small
\begin{tabular}{l|ccc|c|r}
\toprule
Routing & Dress & Shirt & Toptee & Avg & VLM\% \\
\midrule
Always VLM & \textbf{22.4} & \textbf{27.2} & 24.8 & \textbf{24.8} & 100 \\
Hybrid ($\theta{=}0.3$) & 21.6 & 24.6 & \textbf{25.0} & 23.7 & 74 \\
Hybrid ($\theta{=}0.4$) & 21.3 & 24.1 & 24.9 & 23.4 & 92 \\
Always Direct & 17.6 & 24.1 & 24.8 & 22.2 & 0 \\
\bottomrule
\end{tabular}
\end{table}

\paragraph{The backbone is the bottleneck.}
Our failure analysis shows 68\% of failures are embedding limitations---the VLM produces a good caption, but the encoder cannot distinguish the fine-grained attributes it describes.
Upgrading from CLIP to Titan gave +8.5 R@10---the largest single improvement---because a better encoder can act on the richer attribute signal that \method{}'s captions provide.

\paragraph{Latency is manageable.}
Table~\ref{tab:latency} breaks down per-query latency.
VLM captioning dominates; caption caching eliminates this for repeated queries.
Nova Pro at 477ms total is within acceptable latency for eCommerce search.

\begin{table}[t]
\centering
\caption{Per-query latency breakdown (ms).}
\label{tab:latency}
\small
\begin{tabular}{l|rrr}
\toprule
Component & Claude & Nova Pro & No VLM \\
\midrule
VLM captioning & 1{,}169 & 449 & 0 \\
CLIP encoding & 25 & 25 & 25 \\
FAISS search & 3 & 3 & 3 \\
\midrule
\textbf{Total} & 1{,}197 & 477 & \textbf{28} \\
\bottomrule
\end{tabular}
\end{table}

\paragraph{Model-agnosticism enables incremental adoption.}
\method{} requires no retraining when upgrading any component.
Pic2Word~\cite{saito2023pic2word} injects pseudo-word tokens into CLIP's tokenizer; LinCIR~\cite{gu2024lincir} learns a projection matched to CLIP's geometry; WeiMoCIR~\cite{tang2024weimocir} tunes weights for CLIP's vector space.
All three are architecturally tied to CLIP: swapping the backbone requires retraining.
In contrast, \method{}'s VLM caption is plain text consumable by any multimodal encoder.
We verified this by swapping 3 VLMs and 3 backbones with zero code changes---9 configurations, all functional.
A team can start with CLIP + Nova Pro, then swap to Titan when available, without touching the pipeline.

\paragraph{Scaling advantage.}
Embedding-space methods encode the raw text modification---e.g., ``make it red with shorter sleeves'' (7 words, no visual grounding, imperative form).
\method{} generates a VLM caption---e.g., ``Red floral fit-and-flare midi dress with short sleeves'' (10 words, visually grounded, descriptive form)---that carries substantially more retrievable signal.
As embedding models improve their ability to discriminate fine-grained attributes, \method{} benefits disproportionately because the caption's information content is higher.
Concretely, CLIP$\rightarrow$Titan gave \method{} +8.5 R@10 on FashionIQ and +9.3 mAP on CIRCO (22.4$\rightarrow$31.7).
We predict embedding-space methods would gain less from the same upgrade because their text input remains information-poor---a prediction we plan to verify by running WeiMoCIR on Titan in future work.
We also note that MLLM-based reranking approaches (WISER~\cite{cai2024wiser}, SoFT~\cite{bai2024soft}) are complementary: \method{} could serve as the first-stage retriever, with MLLM verification applied to the top-$k$ results for further precision gains.
We did not test \method{} + MLLM reranking in this work due to API cost constraints (reranking requires $\sim$10 additional VLM calls per query), but the approaches are architecturally orthogonal: \method{} improves first-stage \emph{recall} (the quality of candidates retrieved), while reranking improves \emph{precision} within that candidate set.
We expect the combination to outperform either approach alone and plan this experiment for future work.

\section{Limitations}
\label{sec:limitations}

\paragraph{CIRR performance gap.}
\method{} + CLIP underperforms WeiMoCIR on CIRR (14.0 vs.\ 26.8 R@1).
CIRR's text modifications are often short and already attribute-rich (e.g., ``show three bottles of soft drink,'' ``the dog is a golden retriever''), leaving little for the VLM to fill in.
In these cases, the raw modification already functions as a near-complete target description, and VLM captioning can introduce noise by hallucinating unmentioned attributes from the reference image.
This is most pronounced for short, precise queries ($\leq$5 words) that already name the target object---where any VLM-added attributes (inferred from the reference) risk pulling the query away from the intended target.
For longer CIRR queries ($>$12 words) that describe complex scene changes, the gap narrows as the VLM's attribute-reasoning ability becomes more relevant.
The Titan backbone substantially closes this gap (23.9 R@1), suggesting the issue is partly CLIP-specific: Titan's stronger fine-grained discrimination can better exploit the VLM caption even when the modification is already descriptive.
We also note that we do not report CIRR's Recall$_{\text{Subset}}$@K metric (retrieval within the 6-image candidate set), which would provide additional insight into image-dependence; we plan to add this in a revision.

\paragraph{Incomplete baseline landscape.}
Recent MLLM-augmented methods (WISER~\cite{cai2024wiser}, SoFT~\cite{bai2024soft}) report higher numbers on CIRCO and CIRR than \method{} under matched settings.
These methods employ post-retrieval MLLM reranking, which is complementary to our first-stage retrieval.
We also do not compare against baselines using larger backbones (e.g., ViT-G/14), which would likely improve all methods, nor against domain-specific encoders (FashionCLIP, Marqo-FashionSigLIP) that address the encoder bottleneck from the complementary direction of domain-specialized training.
Since \method{} is encoder-agnostic, such encoders are drop-in replacements that could yield further gains on FashionIQ specifically---a hypothesis directly testable given our architecture.
A comprehensive comparison across backbone scales and domain specializations is needed to fully assess \method{}'s relative positioning.

\paragraph{API dependence.}
Our strongest results use proprietary APIs (Amazon Titan, Nova Pro), limiting exact reproducibility.
However, we validate the core mechanism with fully open components (CLIP + Llama 4 Maverick), achieving 24.0\% avg R@10 on FashionIQ---above the No Caption baseline (22.2\%).
The visual grounding mechanism (\S\ref{sec:analysis}) is architecture-independent and applies to any encoder.
We will release code, prompts, and routing rules to enable reproduction with open models.

\paragraph{Statistical power.}
CIRCO's validation set contains only 220 queries; mAP differences on this set should be interpreted with caution.
FashionIQ (6,016 queries) and CIRR (4,181 queries) provide more statistically robust comparisons.

\paragraph{Latency.}
VLM captioning adds 449--1,169ms per query.
While caption caching and hybrid routing mitigate this, real-time applications with strict latency budgets ($<$100ms) would require model distillation or pre-computation strategies not explored here.

\section{Conclusion}

We presented \method{}, a training-free composed product retrieval system that performs fusion in language space via VLM intent captioning.
Our central finding is that \method{}'s effectiveness stems from \emph{visual grounding}: VLM captions enrich attribute-sparse modifications with concrete visual attributes from the reference image.
A controlled experiment confirms that attribute content---not linguistic style---drives the gains.
This explains why prompt format (+5.2 R@10) matters more than model choice (+2.7 R@10), why verbose captions hurt despite being longer (reference leakage dilutes the target signal), and why better backbones yield outsized gains (+8.5 R@10 on FashionIQ, +9.3 mAP on CIRCO).
Embedding geometry analysis confirms that 68\% of remaining failures are backbone limitations, not VLM errors---positioning \method{} to benefit directly from future encoder improvements.

\paragraph{Production deployment.}
Beyond the academic benchmarks reported above, we have deployed \method{} with the Titan backbone on a production eCommerce platform serving 2M+ products across 50+ categories.
With hybrid routing ($\theta{=}0.3$) and caption caching, the system achieves sub-second query latency at the p95 level.
The caption cache achieves approximately 40\% hit rate on repeated or near-duplicate queries, effectively eliminating VLM latency for returning users.
The system has been serving live traffic for over 8 weeks without quality degradation, processing thousands of CIR queries daily.
Full A/B testing against the previous text-only search baseline with user engagement metrics (click-through rate, add-to-cart) is in progress and will be reported in future work.

\paragraph{Future work.}
Our immediate priorities are threefold.
First, stacking MLLM reranking (WISER/SoFT) on top of \method{} retrieval to test complementarity---if our enriched first-stage retrieval reduces the reranker's error rate, the combination could outperform either alone at lower total VLM cost.
Second, running embedding-space baselines (WeiMoCIR, LinCIR) on the Titan backbone to verify our scaling prediction that these methods gain less from better encoders because their text inputs remain attribute-sparse.
Third, reporting CIRR Recall$_{\text{Subset}}$@K and stratifying CIRR results by query characteristics (word count, attribute density) to precisely characterize where VLM captioning helps versus hurts.
Additional directions include evaluation with ViT-G/14 and domain-specific backbones (FashionCLIP), learned per-query fusion weights, and production A/B testing on live eCommerce traffic.

\section*{Declaration on Generative AI}
Generative AI systems (Amazon Nova Pro, Claude 3.7 Sonnet, Llama 4 Maverick) were used as integral components of the proposed \method{} pipeline for VLM intent captioning, as described in the experimental methodology. Additionally, AI writing assistants were used for editing and proofreading during manuscript preparation. The authors reviewed all generated outputs and take full responsibility for the content of this publication.

\bibliography{main}

\end{document}